# Detecting Deepfake-Forged Contents with Separable Convolutional Neural Network and Image Segmentation

Chia-Mu Yu[1], Ching-Tang Chang[1] and Yen-Wu Ti[2]

[1]Department of Computer Science and Engineering, National Chung Hsing University

[2]College of Artificial Intelligence, Yango University

{chiamuyu, justwaybomb, tiyenwu}@gmail.com

Recent advances in AI technology have made the forgery of digital images and videos easier, and it has become significantly more difficult to identify such forgeries. These forgeries, if disseminated with malicious intent, can negatively impact social and political stability, and pose significant ethical and legal challenges as well. Deepfake is a variant of auto-encoders that use deep learning techniques to identify and exchange images of a person's face in a picture or film. Deepfake can result in an erosion of public trust in digital images and videos, which has far-reaching effects on political and social stability. This study therefore proposes a solution for facial forgery detection to determine if a picture or film has ever been processed by Deepfake. The proposed solution reaches detection efficiency by using the recently proposed separable convolutional neural network (CNN) and image segmentation. In addition, this study also examined how different image segmentation methods affect detection results. Finally, the ensemble model is used to improve detection capabilities. Experiment results demonstrated the excellent performance of the proposed solution.

## 1 INTRODUCTION

### 1.1 Image and Video Forgery

With the popularity of consumer electronic products and digital cameras, a large number of digital images and videos have been created. According to statistics, nearly two billion images are added every day on the Internet, and most people can see a large number of images and videos on the Internet every day. This trend has also led to the creation of various digital image editing software to facilitate users in modifying their work. However, such software also provides malicious users with tools to change images. The purpose of digital image forensics research is to identify a forged digital image to avoid damage caused by the image. At present, there are numerous ways to identify image forgeries [10, 26]. Most methods are based on checking the color and pixel details reconstructed by the processing pipeline of the digital camera or determining the presence of alterations in the details of an image. Among these methods, image noise [16] is a useful indicator that can be used to determine whether an image has been clipped. In addition, image compression artifacts [2] are also a useful detection indicator.

The difficulty becomes significantly increased when attempting to detectforged video. Atpresent, up to 100 million hours of video are played on the Internet every day. Therefore, the technology of forging video may cause great damage. Since the compression of a video image will result in the degradation of each frame, the various methods for detecting forged images described in the previous paragraph are insufficient for the detection of forged videos. The main method currently used to detect forged videos is to check for changes in details in a video, however it is difficult to know the edition of the video.

In recent years, deep learning has achieved great results in various applications of digital images, such as Barni et al. [2], whose method is used to detect double JPEG compression. Rao and Ni [25] used deep learning to identify whether digital images had been spliced. Bayar and Stamm [3] used deep learning to identify various forged images. Rahmouni et al.'s method [24] is used to distinguish between photos taken by traditional cameras and digital images generated by computers.

## 1.2 Deepfake

In addition to detecting whether a video has been processed, deep learning is also used to fake videos. One of the best-known methods is a tool called Deepfake, which can be used to forge the expressions and actions of a person's face in a video (see Figure 1). The method used by Deepfake to forge video is similar to Face2Face [2], however the latter uses traditional computer vision techniques to achieve the goal without introducing deep learning. Deepfake, which came out in 2017, can replace the faces of characters in a video with the faces of another person. DeepFake differs from manual image forgery techniques in that it can quickly generate a large number of forged images. If counterfeiters want to manually falsify and output a large number of forged images, they must manually process each image, and must have skilled image editing techniques. Compared to Deepfake, the results of manual forgery can be more realistic, and the traces left are less obvious. However, the technical threshold to achieve this level of forgery is relatively high. Using DeepFake, a counterfeiter can easily make forged video as long as enough training data is provided, without any professional forgery or editing skills.

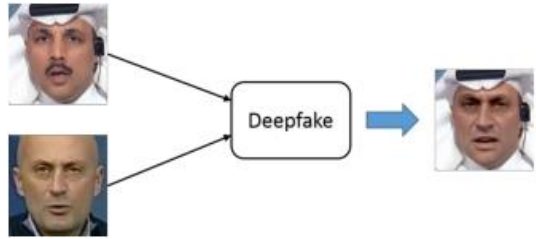

Fig. 1. Two real pictures are inputted into the Deepfake model to obtain a forged image

Similarly, in order to detect a film or image generated by Deepfake, an AI model must learn the features of facial forgery. The AI model must thus learn the correlation between large amounts of data, and after a lot of training, the AI model will be able to determine the optimal weight distribution. As a result, with increasing amounts of data, such detection techniques have very high hardware requirements. However, forged images for training can be easily obtained on the Internet.

## 1.3 Contribution and Organization

This paper proposes a method for detecting facial forgery. The proposed method segments images as patches, and each patch identifies subtle features by convolutional neural networks. Based on these features, the categories of patches can be determined, and one can determine whether the image is a fake by considering the categories of all patches.

The organization of this study is as follows. Section 1 briefly introduces the tools used to forge video and its principles and applications. Section 2 evaluates the state-of-the-art face forgery and detection. Section 3 details our proposed network architecture. Section 4 discusses the datasets and performance evaluations. Also, our proposed method compares the accuracy of detection with the state-of-the-art. The receiver operating characteristic (ROC) curve is used as a benchmark for analysis that can know the classification capability of the model in binary categories. Finally, Section 5 is the conclusion and prospect of this paper.

## 2 RELATED WORK

### 2.1 Digital Manipulation

The forgery of digital media has become more widespread. For example, Ren et al. [27] applied unsupervised learning to simulate realistic sounds using a small amount of sound training data. The method proposed by Fried et al.[12] allows a user to arbitrarily modify what a person in a video says. These methods work very well when forging voice signals, and if used to attack a voice-controlled system, pose a security risk. Now, research results are able to not only forge voice, but also digital images. The controversial DeepNude [8] is a well-known example, which converts images of a clothed person into nude images using the Pix2Pix algorithm. Many related studies have been conducted on facial forgery, and four common methods of facial forgery have been identified, including DeepFake [7], FaceSwap [9], Face2Face [29], DeepFaceLab [6]. These are automatic face forgery methods, and this type of technology is also collectively known as Deepfakes [13]. Recently, this technology has had a major negative impact. This study therefore addresses the following four major methods for face forgery.

DeepFake[7]: Deepfake's idea is to train two autoencoders in parallel. Its structure can be adjusted according to the user's resources and the demand for output image quality. The process of Deepfake is as follows: Use the faces of two different people, A and B, and then train autoencoders to use the A and B face image data sets to reconstruct the faces of A and B respectively. Separate the decoders used for the images of A and B, but share the weights of the encoding parts of the two autoencoders. Optimize the autoencoders, and then any image (including image A) can be encoded by the above-mentioned shared encoder and then decoded by the decoder used to process image B. Therefore, this method requires an encoder that encodes information such as the illuminance, position, and expression of the face, and then trains a decoder for each person to reconstruct the features, shapes, and other details of the face. Contextual information and morphological information may have to be handled separately.

Deepfake works very well and has become very popular. After the aforementioned tools are prepared, the final steps are to get the target video, find the target face of each frame in the video, and align them. The aforementioned autoencoders are then used to generate a fake face and use it to replace the original face in the video.

However, this type of technology is not perfect, because the above steps may fail, especially when the face that appears in the video is partially obscured. Whether it is the process of finding the original face or inserting the forged face, errors may occur and large blurred images may appear. In theory, increasing computing resources to use more complex networks can increase the success rate. On top of that, the choice of the size of the encoding space is also an interesting topic. In the case of the input compression, if the encoding space is too small, it will cause difficulties in making a detailed reconstruction of the face and create a blurry result. On the other hand, if the coding space is too large, the details of the reconstruction can be improved; however, the morphological data will be passed to the decoder, thus causing the forged face to approximate the input face, which is contrary to the target of the fake face.

In addition to the above, reenactment methods are also approaches for converting the facial expressions of a target into the facial expressions of another image [x9]. At present, the best of these methods is Face2Face proposed by Thies et al. It can perform markerless facial reenactment in real time, and the effect is realistic. This method takes a few minutes of the target video to build the face model. When using Face2Face, the face in the source and target videos should be tracked simultaneously. The face of the target can then be superimposed with the modified face shape to form the facial expression in the source video.

FaceSwap [9]: This method detects 3D-based facial landmarks to obtain a target's face shape, and converts the face area of the source image to that of a target image using a 3D model. The locations of facial landmarks are obtained using the technology proposed by Dlib[18], and the 3D facial template is provided by CANDIDE. Specifically, Gauss Newton fine-tunes facial landmarks and 3D templates are used to create accurate 3D models and to convert source faces to target faces. This method performs fast face conversion using only the CPU. The model is projected into the image space using the following equation (1):

$$s = \alpha P(S_0 + \sum_{i=1}^{n} w_i \times S_i) + t \qquad (1)$$

where s is the projected shape, $\alpha$ is the scaling parameter, P is the rotation matrix which rotates the 3D face shape, $S_0$ is the neutral face shape, $w_i$ are the blendshape weights, $S_i$ are the blendshapes, t is a 2D translation vector and n is the number of blendshapes.

Face2Face [29]: Face2face is a facial reenactment method that converts facial expressions in a video into the facial expressions of the user. In order to make the face and shadow of each frame the same, it uses the dense photometric consistency measure. The method establishes a 3D facial model in the first frame, and modifies the facial expression of the frame after the 3D model is reconstructed to implement instant video facial reconstruction. Most existing real-time face trackers are based on sparse features. Face2Face uses all the information detected in the input. This method is called a dense face tracker. However, Face2Face determines the parameters of the face model by minimizing the error between the original image and the composite face image. The resulting composite model is very close to the input, so it is difficult to distinguish between synthetic faces and real faces.

DeepFaceLab[6]: DeepFaceLab is a variant of the original DeepFake. To improve the blurry regions in the generated image, DeepFaceLab replaces the autoencoder with GAN as the architecture of the training network. Additional training of a set of attention masks helps to handle occlusion and produces natural skin tone. Improved eye orientation is added in the loss function to make the fake image more realistic and consistent with the input face. In addition, many parameters can be adjusted to increase training effectiveness. For example, they can blur the boundaries of the mask, increasing the area of the source face to the target face, etc. These fine-tuned parameters can make the fake face more realistic.

## 2.2 Digital Forensics

Traditionally, in order to secure against replay attacks on access control systems, researches have applied local binary pattern (LBP) face anti-spoofing [4, 30, 31]. These techniques mainly detect whether an attacker is attempting to confuse an identification system with photos or masks. However, the synthetic images considered in this study are more challenging to identify than traditional attacks. Now, digital forgeries can not only learn the feature distribution of real images, but can also generate images that are more difficult to distinguish. Although the synthesized image is highly similar to the actual picture, there are several methods for detecting such synthesized images. Zhou et al. [32, 33] proposed a two-stream network to detect image manipulation. Li et al.[20] used long short-term memory (LSTM) to detect abnormal blinks. This study uses state-of-the-art methods as the objects for comparison. Two baseline approaches used in this study are briefly described below.

Mesonet [1]: Mesonet is a simple design of CNN architecture for forgery detection. Although network only uses 27,977 training parameters, its detection capability is highly accurate. Mesonet was run smoothly in this study without the need for a powerful computer. Table 1 shows the Mesonet model architecture.

Capsule-Forensics [23]: Nguyen et al. proposed Capsule-Forensics to detect forged images. This architecture consists of two steps. In the first step, the face of the character in the video is detected and scaled to $128 \times 128$. Some of the VGG network [28] are used to extract latent features for input into the capsule network. Specifically, the third maxpooling layer is obtained before the relu layer because the size of the capsule network input must be reduced. In the second step, the network consists of three primary capsules and two output capsules, which are used to detect real and false images, respectively. The latent features extracted by a part of the VGG network are input to the three primary capsules. The output of the three capsules ($u_{j|i}$) are dynamically routed to the output capsule ($v_j$) for r iterations. The Capsule-Forensics network has approximately 2.8 million parameters, which is a relatively small number compared to other similar networks. In addition, Gaussian random noise is added to the 3D weight tensor $W$, and additional compression is applied before routing iteration (2).

$$s_j = squash(s_j) = \frac{\left\| s_j \right\|^2}{1 + \left\| s_j \right\|^2} \frac{s_j^2}{\left\| s_j \right\|} \tag{2}$$

Specifically, the squash function shrinks the $s_j$ vector from 0 to unit length. Increased noise helps reduce overfitting, while additional equations make the network more stable.

| Layer name | Output shape | Kenel | Activation | Previous layer |
|---|---|---|---|---|
| Input | $256 \times 256 \times 3$ | | | |
| Conv2D_1 | $256 \times 256 \times 8$ | (3,3) | relu | Input |
| BatchNorm_1 | $256 \times 256 \times 8$ | | | Conv2D_1 |
| MaxPooling_1 | $128 \times 128 \times 8$ | (2,2) | | BatchNorm_1 |
| Conv2D_2 | $128 \times 128 \times 8$ | (5,5) | relu | MaxPooling_1 |
| BatchNorm_2 | $128 \times 128 \times 8$ | | | Conv2D_2 |
| MaxPooling_2 | $64 \times 64 \times 8$ | (2,2) | | BatchNorm_2 |
| Conv2D_3 | $64 \times 64 \times 16$ | (5,5) | relu | MaxPooling_2 |
| BatchNorm_3 | $64 \times 64 \times 16$ | | | Conv2D_3 |
| MaxPooling_3 | $32 \times 32 \times 16$ | (2,2) | | BatchNorm_3 |
| Conv2D_4 | $32 \times 32 \times 16$ | (5,5) | relu | MaxPooling_3 |
| BatchNorm_4 | $32 \times 32 \times 16$ | | | Conv2D_4 |
| MaxPooling_4 | $8 \times 8 \times 16$ | (4,4) | | BatchNorm_4 |
| FullyConnected_1 | 1024 | | relu | MaxPooling_4 |
| FullyConnected_2 | 16 | | Leakyrelu | FullyConnected_1 |
| FullyConnected_3 | 2 | | softmax | FullyConnected_2 |

Table 1. The network architecture of Mesonet. The maximum tensor channel is only 16, and the multi-layer pooling is used to reduce the training size

## 3  THE PROPOSED METHOD

The proposed detection forgery scheme can be divided into two steps, as shown in Figure 2. The first step is to extract the face area of each frame in the video. These facial images are acquired by alignment and cropping processing, and input to the classifier for processing. The second step divides the blocks of the pre-processed face area and trains them using CNN. Finally, hard voting is used to determine the label of the image. The proposed method of image pre-processing is different from most approaches. It does not just use a single processing step, but combines the segment image and the overall image for feature analysis to improve accuracy. The network architecture we proposed is a separable CNN, which is different from the fully connected CNN used in previous research on detecting fake videos. The architecture has added some functional layers for improvement accuracy. We describe the relevant details in subsection 3.3. The architecture has added some functional layers for improvement accuracy.Then the proposed method can achieve high accuracy.

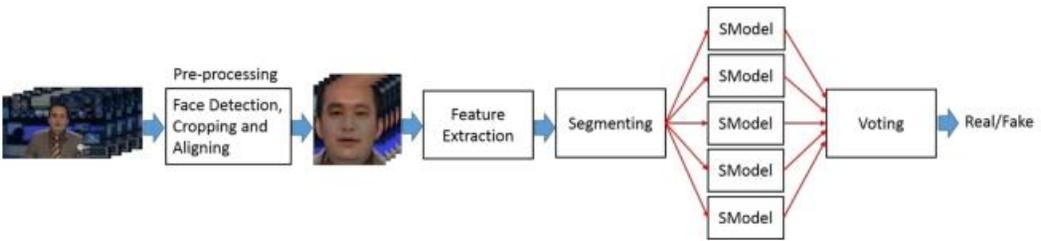

Fig. 2. Overview of the proposed architecture. The facial image features are extracted to obtain latent features (Figure 3). SModel is a small CNN network that classifies blocks, as shown in Figure 5.

## 3.1 Image Pre-Processing

This study extracts each frame of the video, and then obtains the facial landmarks, in order to give the faces of different individuals similar distributions in the same block. Each extracted face is aligned so that the eyes are on the same horizontal line, and saved as a $256 \times 256$ pixel image. This processing method allows each face area to be in the same coordinate position for better results.

## 3.2 Facial Forgery Detection

As mentioned earlier, the proposed method extracts features from the entire image based on CNN, as shown in Figure 3. In order to enable the model to learn fine-grained features, this study segments potential features and trains different block features. The latent features are cut horizontally and vertically to form four blocks of the same size. In Figure 4, $s_1$ to $s_5$ represent the number of each segment block, and five latent features are trained by the CNN network, respectively. Each block can determine whether the feature is real or fake, as shown in Figure 5. Finally, the category of the image is counted by voting method.

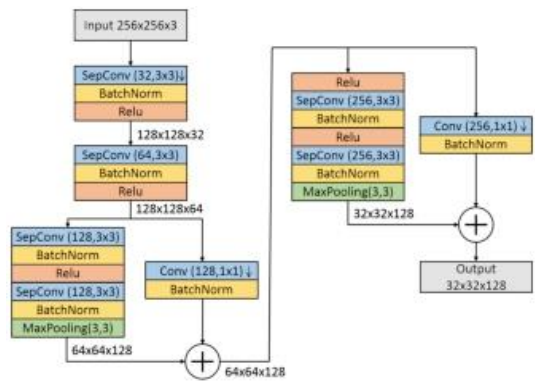

Fig. 3. The network used to extract latent features

## 3.3 Network Architecture

In the proposed network architecture, the input is a $256 \times 256 \times 3$ pixel image. The tensor is obtained by a convolution network with a length and width of 32, and a depth of 128. The tensor is divided into four equal parts, and the original latent features are retained. Therefore, four $16 \times 16 \times 128$ and one $32 \times 32 \times 128$ tensors are obtained. These tensors are trained through the same architecture network. The softmax layer calculates the probability of each label for a real and fake classification.

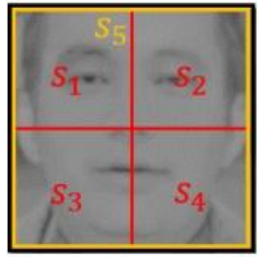

Fig. 4. Each number is an image block feature, and the orange rectangle represents the full feature. Our method divides the feature into four segments of the same size and one entire latent feature.

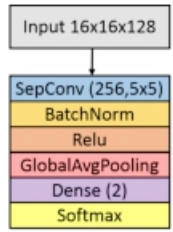

Fig. 5. SModel is an image classifier, which classifies categories according to image features.

The proposed method replaces the original fully connected CNN with a separable CNN [5] to effectively reduce the parameters required for training. The batch normalization layer [15] and the rectifying linear units are added after each convolutional layer, and the model generalization ability is improved with the Resnet architecture [14]. In addition, the global average pooling layer [21] replaces the fully connected layer to improve accuracy.

The separable CNN, illustrated in Figure 6, differs from the original CNN in that it is divided into two steps. First, the weight of the input tensor is calculated by a $1 \times 1$ convolution. This step increases the number of channels for the tensor. Next, each channel uses an $N \times N$ convolution kernel to process the corresponding tensor using weights. This method can be used to reduce the number of parameters required without significantly reducing the accuracy.

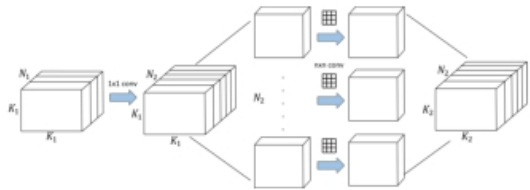

Fig. 6. Separable CNN is a variant CNN that reduces the number of parameters in training.

## 4 EXPERIMENTAL RESULTS

The experiments conducted in this study used the FaceForensics++ (FF ++) data set. In addition, this studyusedthereceiveroperatingcharacteristic(ROC)curvetomoreflexiblypredicttheobservation

probability belonging to each class of classification problems, rather than directly predicting the class. This helps explain the state of the model in binary classification prediction. This study uses FFmpeg[11] to extract each frame of the video, and then obtains the facial landmarks by the method proposed by Dlib[18]. This study used ADAM [19] to optimize and set $\beta_1 = 0.9$, $\beta_2 = 0.999$ for the initial learning rate, and $10^{-6}$ for decay. Cross entropy was used as a loss function, which is often used when the activation function is softmax. The number of epochs was set to 200, and batch size was 32. Each epoch was used to train the entire data set.

## 4.1 Dataset

FaceForensics++ includes 1000 original videos, as well as versions processed by DeepFake, Face2Face and FaceSwap. Each video extracts only a few frames as a data set. This study mixed the three fake videos together as a negative sample of training and verification. A total of 18,128 images were used for training, and 2,647 images were used for verification. 4,059 images were collected on a video sharing platform, and DeepFaceLab processing was used as a negative sample, as shown in Table 2 in order to test the robustness of images generated based on GAN. The experiments also tested the StyleGAN [17] scheme proposed, which uses GAN to generate 1024 × 1024 pixel face images that look very similar to real photos. However, the images generated by StyleGAN do not replace the faces of the characters in the original images, but only use GANs to generate faces, so strictly speaking, it is not the same as the method of replacing characters' faces such as Deepfake. Figure 7 shows face images produced by various methods.

| Dataset | Training | Validation | DeepFaceLab | StyleGAN |
|---------|----------|------------|-------------|----------|
| Positive | 7,650 | 1,002 | 1,439 | 1,439 |
| Negative | 10,478 | 1,645 | 2,620 | 2,000 |

Table 2. Training and validation data are a mixture of FaceForensics++. DeepFaceLab and StyleGAN were used as testing data, and they used the same positive samples.

## 4.2 Evaluation Metrics

The following will outline the ROC evaluation metrics and t-SNE [22] of data visualization. The receiver operating characteristic (ROC) curve presents the performance of a binary classifier system in a specific classification or discrimination threshold. The y-axis of the graph is the true-positive rate (TPR), which is the sensitivity. The x-axis is the false-positive rate (FPR), denoted as 1 − specificity. Sensitivity and specificity represent the probability of positive and negative results, respectively. If a cut-off point is suitable for distinguishing between positive and negative effects, it can be used as the optimal threshold for the classifier. Overall, TPR and FPR can be defined as:

$$\text{TPR} = \text{TruePositive}/(\text{TruePositive} + \text{FalseNeɡative})$$

$$\text{FPR} = \text{FalsePositive}/(\text{FalsePositive} + \text{TrueNeɡative})$$

(3)

In this study, when the experiment results are interpreted, the diagonal is used as a reference line. If the ROC curve of the classifier falls on the diagonal reference line, it means that the classifier has no discriminability for that category. If the ROC curve is closer to the upper left, it means that the classifier is more sensitive to classification, and the false positive rate is lower. In other words, the nearest point (0,1) is the cut-off point with the minimum misclassification, while the sensitivity is the maximum, and the false-positive rate is the minimum.

The area under curve (AUC) is defined as the area under the ROC curve, which can be used to evaluate the accuracy of the classifier. Specifically, AUC is a probability value. When a positive

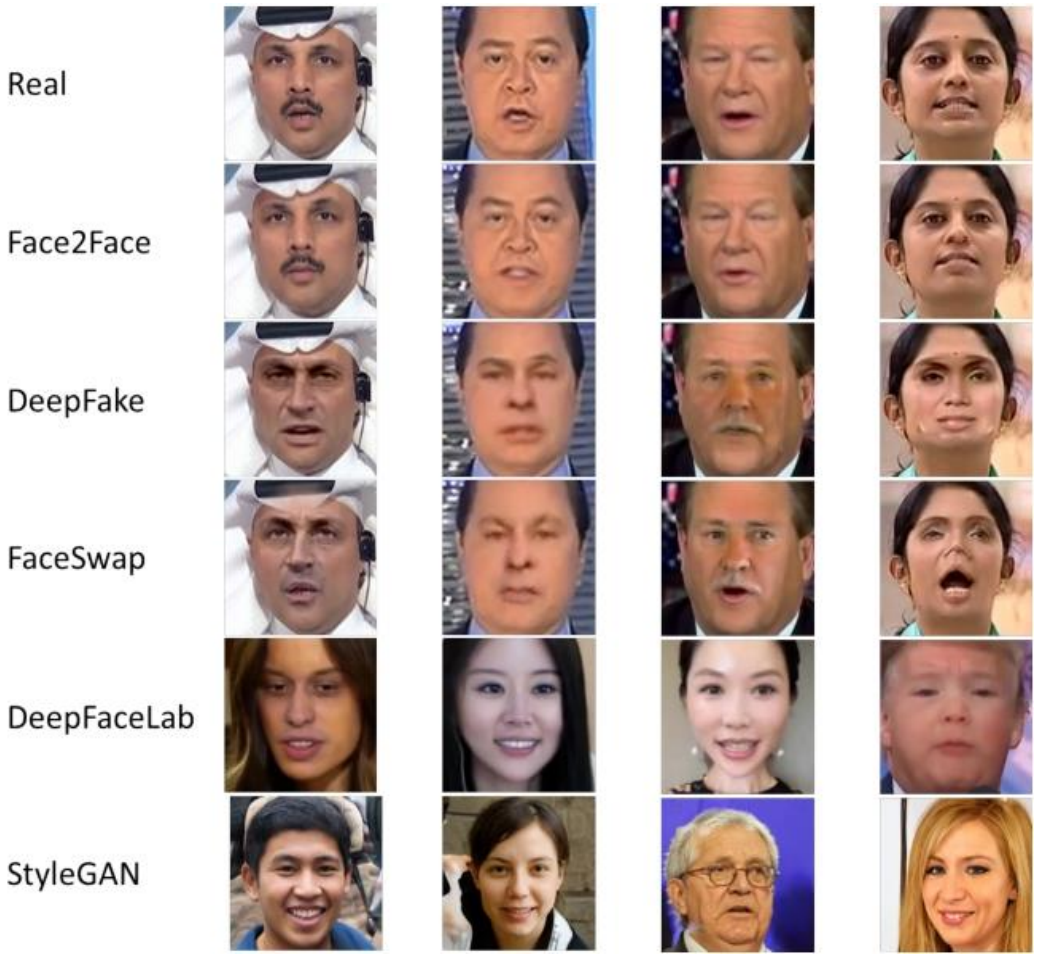

Fig. 7. Each row corresponds to a method of generating an image. The image of the row corresponding to DeepFaceLab is collected from a video sharing platform. The row corresponding to StyleGAN is provided by the NVIDIA team. StyleGAN is a method proposed by the NVIDIA team, and the corresponding row is the image generated by StyleGAN

sample and a negative sample are randomly chosen from the data, the probability that the classifier will rank the positive sample before the negative sample based on the calculated score value is the AUC value. Naturally, a larger AUC value means that two categories can be effectively distinguished.

The ROC curve has excellent properties in use. In the actual processed data, the nature of the data is usually unbalanced. That is, there are far more negative samples than positive samples (or vice versa). Moreover, the distribution of positive and negative samples in the test data may also change over time. However, this study uses fixed data sets, so the data does not change over time.

T -distributed stochastic neighbor embedding (t-SNE) is a visual machine learning algorithm. It is a nonlinear dimensionality reduction technique that compresses high dimensional data into two- or three-dimensional spaces for visualization. This method is very helpful when observing high-dimensional data. It allows us to visually understand some characteristics of the data. The t-SNE algorithm consists of two main phases. First, SNE constructs a probability distribution between high-dimensional objects such that similar samples have a higher probability of being selected, while dissimilar samples have a lower probability of being selected. Second, SNE constructs the probability distribution of these samples in a low-dimensional space, making the two probability distributions as similar as possible. The algorithm uses Euclidean distance as the basis for the similarity measure. Therefore, clusters between datasets can be obtained by t-SNE.

In this paper, we use t-SNE to observe the feature distribution of the dataset and perform a deeper comparison between the performance results of different methods. Figure 8 shows the feature distribution of the FaceForensics++ data set by t-SNE. It can be observed that DeepFake and Face2Face have similar feature distributions, so the generalization effect is better than that of FaceSwap. However, according to our experimental results, despite the difference in feature distribution, our proposed methods can achieve excellent recognition accuracy, highlighting the superiority of the proposed method in this paper. The relevant details will be presented in the following sections.

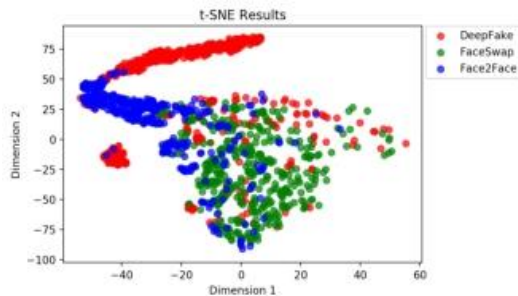

Fig. 8. The FaceForensics++ dataset projects images into 2-dimensional space by t-SNE and visualizes Deepfake, FaceSwap and Face2Face images in 2-dimensional space.

### 4.3 Comparison of State-of-the-Art Methods

To ensure the fairness of the results, this study trained and tested with the same parameters in all models. As shown in Figure 9, different methods were trained by the FaceForensics++ data set. Each method exhibited high accuracy in the validation data of the same feature distribution in 9a to 9d. However, in the new dataset, the accuracy of the model classification was worse because the feature distribution of training and testing data were different. The final accuracy is shown in Table 3. It can be seen that the proposed method demonstrated higher accuracy and AUC values in most of the comparative experiments.

In the FS/DF and F2F/FS rows, the proposed scheme is less accurate than Mesonet. However, with reference to the AUC value, the high accuracy of Mesonet is caused by data imbalance, and there is actually no difference in the classification capabilities of the two models. Generalization of only partial data sets was also tested, as shown in Figures 10 to 15.

Comparing Figure 9f with Figure 9h, the proposed method had a higher classification rate than Capsule on the face-forged DeepFaceLab dataset. According to Figure 9g, for StyleGAN images, when the epoch number is greater than 75, Capsule's detection effect is better than other methods, because when StyleGAN is detected the other methods classify more fake images as positive samples, resulting in lower accuracy results. However, the ROC curve shown in Figure 9h clearly indicates that the proposed model is still able to classify StyleGAN images. From Table 3, the overall advantages of the proposed method can be seen. Note, also, that since Mesonet is composed of a

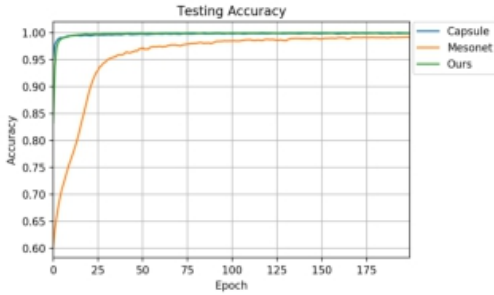

(a) Accuracy on Training data

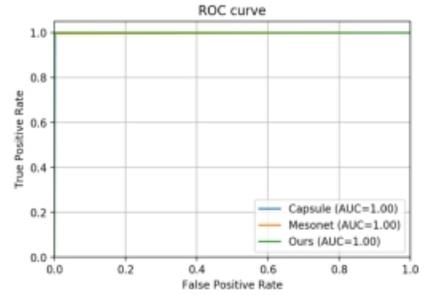

(b) ROC curve on Training data

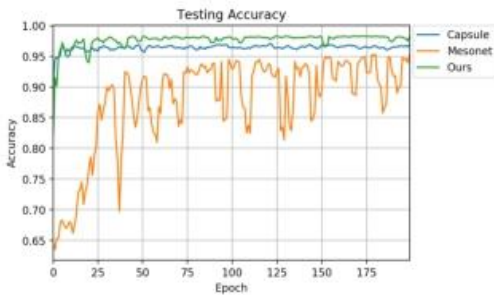

(c) Accuracy on Validation data

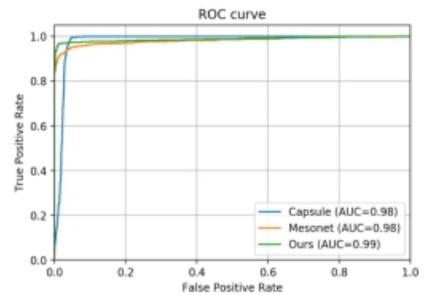

(d) ROC curve on Validation data

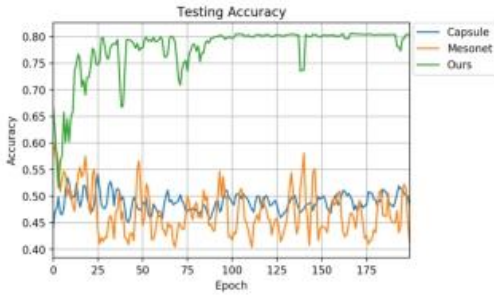

(e) Accuracy on DeepFaceLab

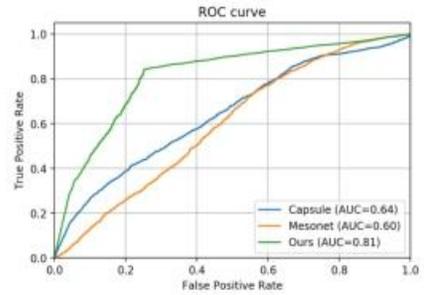

(f) ROC curve on DeepFaceLab

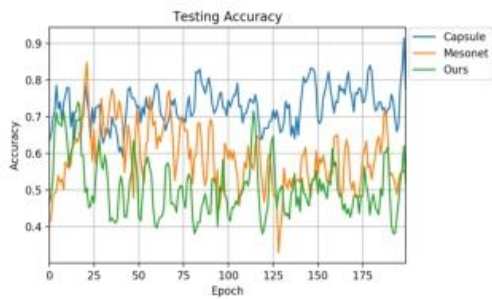

(g) Accuracy on StyleGAN

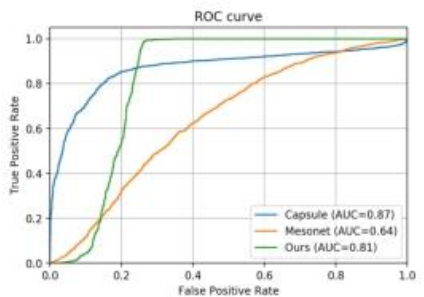

(h) ROC curve on StyleGAN

Fig. 9. Comparison of the accuracy and AUC results of multiple datasets according to various proposed CNN architectures and other state of the art methods

| Training Data | Testing Data | Accuracy (%) | | | Area under the Curve of ROC | | |
|---|---|---|---|---|---|---|---|
| | | Mesonet | Capsule | Ours | Mesonet | Capsule | Ours |
| FF++ | FF++ (Training) | 99.4 | 99.8 | 99.9 | 1.00 | 1.00 | 1.00 |
| FF++ | FF++ (Testing) | 95.1 | 96.9 | 98.1 | 0.98 | 0.98 | 0.99 |
| FF++ | DeepFaceLab | 40.9 | 48.4 | 80.5 | 0.60 | 0.56 | 0.80 |
| FF++ | StyleGAN | 63.9 | 68.2 | 84.9 | 0.56 | 0.87 | 0.81 |
| DF | DF | 99.8 | 99.9 | 99.9 | 1.00 | 1.00 | 1.00 |
| DF | FS | 41.4 | 54.1 | 58.2 | 0.30 | 0.34 | 0.50 |
| DF | F2F | 37.0 | 68.5 | 66.6 | 0.80 | 0.95 | 0.93 |
| FS | FS | 98.8 | 99.6 | 99.6 | 1.00 | 1.00 | 1.00 |
| FS | DF | 55.4 | 50.1 | 51.3 | 0.32 | 0.52 | 0.74 |
| FS | F2F | 57.9 | 51.9 | 78.2 | 0.61 | 0.70 | 0.97 |
| F2F | F2F | 99.6 | 99.6 | 99.8 | 1.00 | 1.00 | 1.00 |
| F2F | DF | 35.8 | 58.1 | 63.3 | 0.88 | 0.95 | 0.99 |
| F2F | FS | 58.8 | 54.1 | 50.8 | 0.51 | 0.52 | 0.80 |
| DF&FS | DF&FS | 98.6 | 99.7 | 99.9 | 1.00 | 0.99 | 1.00 |
| DF&FS | F2F | 37.3 | 70.0 | 92.3 | 0.92 | 0.92 | 0.96 |
| DF&F2F | DF&F2F | 99.3 | 99.6 | 99.6 | 1.00 | 0.99 | 1.00 |
| DF&F2F | FS | 44.5 | 54.2 | 58.6 | 0.30 | 0.49 | 0.74 |
| FS&F2F | FS&F2F | 98.8 | 99.7 | 99.9 | 1.00 | 0.99 | 1.00 |
| FS&F2F | DF | 40.1 | 62.8 | 84.5 | 0.77 | 0.90 | 0.98 |

Table 3. Accuracy (%) of state-of-the-art methods.

simple network architecture, it had difficulty distinguishing between positive and negative samples in the new data set.

Figure 10 shows the experiment results using DeepFake as the training data set. Figure 10c and Figure 10d show that due to the huge difference in feature distribution between DeepFake and FaceSwap, the experiment results of each model only showed lower accuracy and AUC value, which means that the image classification ability was very low. Figure 10f and Figure 12f show that because DeepFake and Face2Face are similar in feature distribution, higher accuracy and AUC results were obtained, indicating that the proposed model had better classification ability. In Figure 10f, the proposed method takes the truncation of the ROC curve as the new classification threshold (threshold = 0.9), and its accuracy increased from 66% to 86%.

From Figure 11c and Figure 11d, although the accuracy of Mesonet was high, the ROC curve indicated that this was due to data imbalance. The same situation also occurs in Figures 12e and Figure 12f. The proposed approach has the ability to group data slightly higher than Capsule. From Figure 13 and Figure 15, since the data distribution of Deepfake and Face2Face is similar, training and testing can obtain better results. However, the distribution of FaceSwap differs significantly from other datasets, resulting in poor experiment results, as shown in Figure 14.

## 4.4 Comparison of Different Segment Methods

This subsection compares the impact of different block slices on accuracy. The experiment tested the latent features of the eight slicing methods. Figure 4 shows the method of cutting images when using five block segments, which is also an optimization method. Figure 16 presents the other approaches we took in our experiments. As shown in Figure 4, the red line denotes that the latent features were sliced horizontally or vertically, with an orange rectangle denoting the central feature.

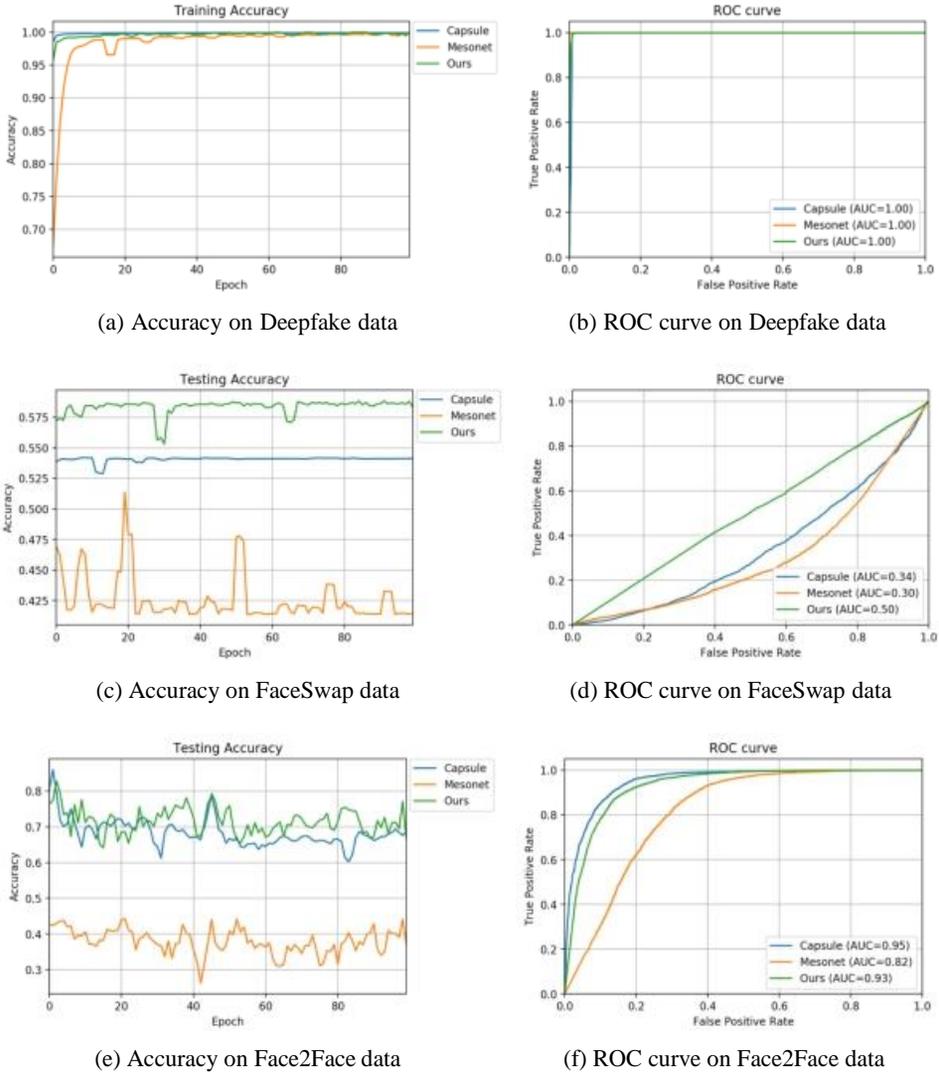

Fig. 10. Training the Deepfake dataset and comparing the accuracy of other datasets by AUC values

In order to maintain the dependent features of the original image, this study used horizontal and vertical slicing of the fixed area instead of randomly picking the block locations. The average segmentation of image features ensures each block has the same shape in order to train the subtle features of the whole image. In addition, based on the Mesonet architecture, the experiment also compared the results obtained using slices and non-slices. This study took a network architecture with a fixed position horizontal cut as an example, as shown in Figure 16. Unlike the original Mesonet, the segmentation network divides the latent features into three equal parts, and uses the voting results as a basis for classification.

In order to maintain the dependent features of the original image, we used horizontal and vertical cutting of the fixed area instead of randomly picking the location of blocks. The average segmentation of image features allows each block the same shape to train the subtle features of the whole image. In addition, results are compared between slice and non-slice — the Mesonet architecture as a baseline. We take the network architecture of horizontally cut at a fixed location as an example, as shown in Table 4. Different from the original Mesonet, the segmented network divides the latent feature into three equal parts and takes voting as the result of classification.

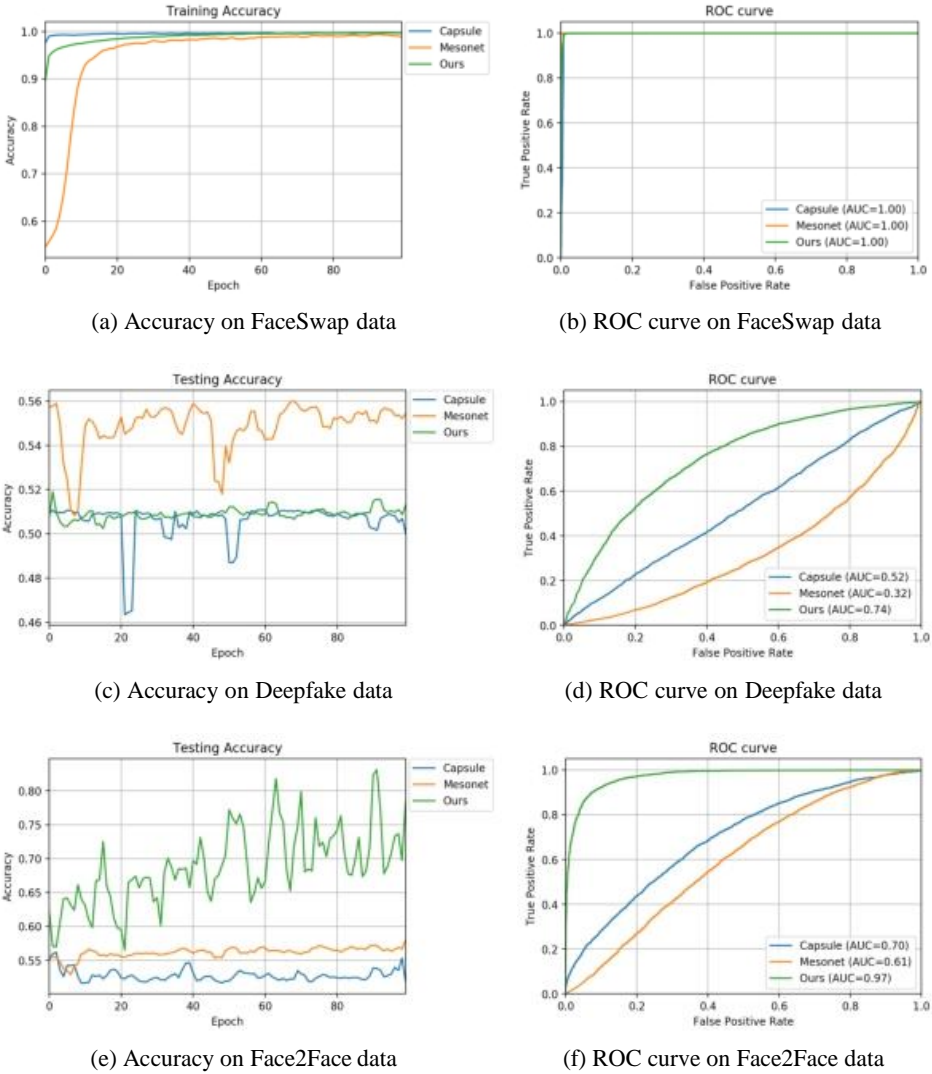

(a) Accuracy on FaceSwap data

(b) ROC curve on FaceSwap data

(c) Accuracy on Deepfake data

(d) ROC curve on Deepfake data

(e) Accuracy on Face2Face data

(f) ROC curve on Face2Face data

Fig. 11. Training on FaceSwap data and comparing the accuracy of other datasets with AUC results.

If the number of horizontal slices is set as x, then this setting is called v$x_h$ . Similarly, if the number of vertical slices is set as y, then this setting is called v$y$. The experiment tested DeepFaceLab and StyleGAN data according to the experiment setup in described Section 4.3. As shown in Table 5, if the input is divided into more block segmentations, the accuracy increases. However, when set to cut into five block segmentations, the accuracy is almost the same as the AUC value, or better. Therefore, the accuracy and AUC values obtained by selecting five segments are the optimal solution.

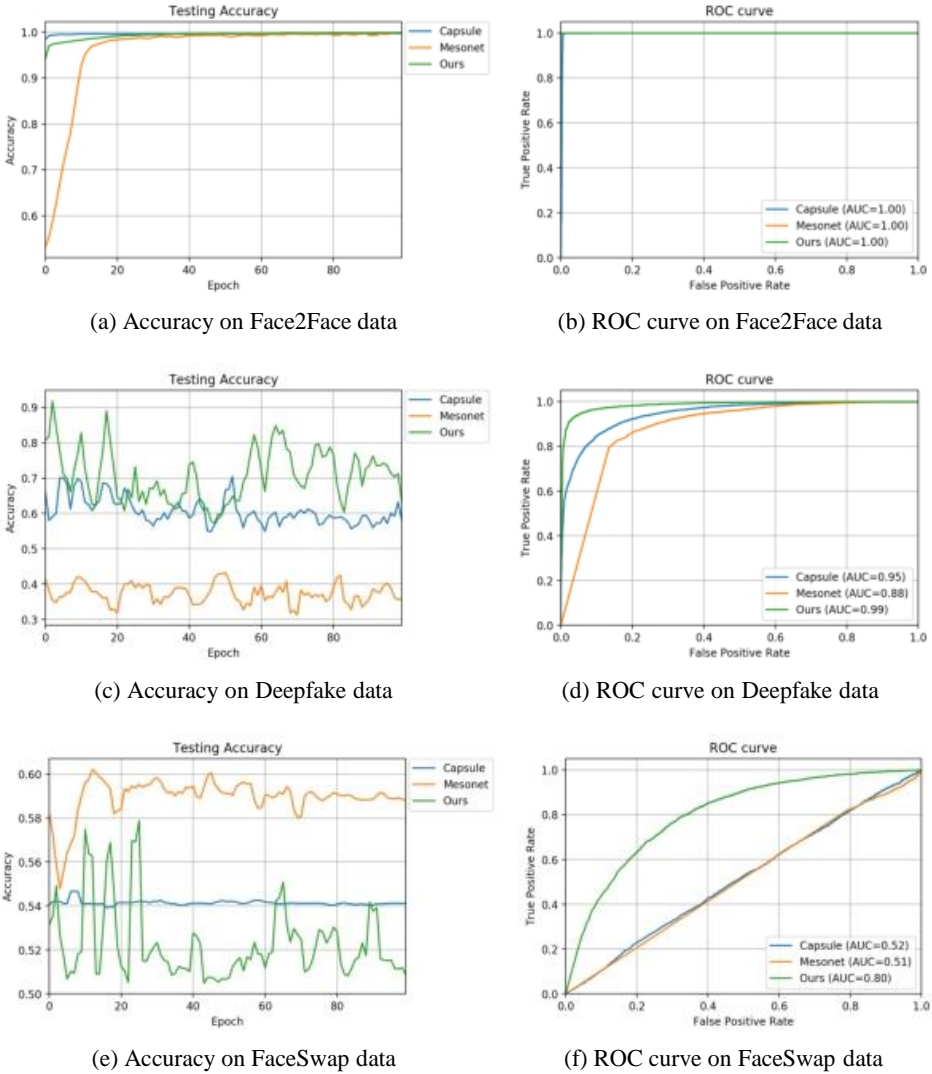

(a) Accuracy on Face2Face data

(b) ROC curve on Face2Face data

(c) Accuracy on Deepfake data

(d) ROC curve on Deepfake data

(e) Accuracy on FaceSwap data

(f) ROC curve on FaceSwap data

Fig. 12. Training on Face2Face data and comparing the accuracy of other datasets with AUC results.

Figure 17 compares the AUC values of each splitting method for the FaceForensics++ dataset. In the dataset of the same feature distribution, the number of image splits is proportional to the result. From Table 5 and Figure 17, we know that the methods of v3_h and v3_v lack the assistance of the central region and the correlation between the features of each region cannot be used to improve accuracy, resulting in poor accuracy. On the other hand, as can be seen from Figure 17c and Figure 17d, increasing the number of image block segments in the untrained data set does not improve the experiment results, but the effect is higher than the uncut method. When 17 block segments are used, the accuracy and AUC values are significantly reduced. If 26 or 37 block segments are used, the model loses discriminatory power. This is because the block segmentation is too thin in this setting, so most block segments will misidentify the label. Hence, cutting the image into too many blocks interferes with feature extraction, and even with the assistance of the central area, the accuracy will remain poor. Therefore, five, seven and ten block segments are the three methods with the highest accuracy. These methods have different performances in different data sets. Overall, the advantages of five block segments are obvious.

| Layer name | Output shape | Kenel | Activation | Previous layer |
|---|---|---|---|---|
| Input | $256 \times 256 \times 3$ | | | |
| Conv2D_1 | $256 \times 256 \times 8$ | (3,3) | relu | Input |
| BatchNorm_1 | $256 \times 256 \times 8$ | | | Conv2D_1 |
| MaxPooling_1 | $128 \times 128 \times 8$ | (2,2) | | BatchNorm_1 |
| Conv2D_2 | $128 \times 128 \times 8$ | (5,5) | relu | MaxPooling_1 |
| BatchNorm_2 | $128 \times 128 \times 8$ | | | Conv2D_2 |
| MaxPooling_2 | $64 \times 64 \times 8$ | (2,2) | | BatchNorm_2 |
| Conv2D_3 | $64 \times 64 \times 16$ | (5,5) | relu | MaxPooling_2 |
| BatchNorm_3 | $64 \times 64 \times 16$ | | | Conv2D_3 |
| MaxPooling_3 | $32 \times 32 \times 16$ | (2,2) | | BatchNorm_3 |
| Conv2D_4 | $32 \times 32 \times 16$ | (5,5) | relu | MaxPooling_3 |
| BatchNorm_4 | $32 \times 32 \times 16$ | | | Conv2D_4 |
| Segmentlayer_1 | $10 \times 32 \times 16$ | | | BatchNorm_4 |
| Segmentlayer_2 | $11 \times 32 \times 16$ | | | BatchNorm_4 |
| Segmentlayer_3 | $11 \times 32 \times 16$ | | | BatchNorm_4 |
| SeparableConv2D_1 | $10 \times 32 \times 256$ | (5,5) | relu | Segmentlayer_1 |
| SeparableConv2D_2 | $11 \times 32 \times 256$ | (5,5) | relu | Segmentlayer_2 |
| SeparableConv2D_3 | $11 \times 32 \times 256$ | (5,5) | relu | Segmentlayer_3 |
| BatchNorm_5 | $10 \times 32 \times 256$ | | | SeparableConv2D_1 |
| BatchNorm_6 | $11 \times 32 \times 256$ | | | SeparableConv2D_2 |
| BatchNorm_7 | $11 \times 32 \times 256$ | | | SeparableConv2D_3 |
| GlobalAvgPool_1 | 256 | | | BatchNorm_5 |
| GlobalAvgPool_2 | 256 | | | BatchNorm_6 |
| GlobalAvgPool_3 | 256 | | | BatchNorm_7 |
| FullyConnected_1 | 2 | | softmax | GlobalAvgPool_1 |
| FullyConnected_2 | 2 | | softmax | GlobalAvgPool_2 |
| FullyConnected_3 | 2 | | softmax | GlobalAvgPool_3 |

Table 4. This example uses the latent features extracted by Mesonet, which is split into three blocks by horizontally. (v3_h)

| | FF++ (vali) | | DeepFaceLab | | StyleGAN | |
|---|---|---|---|---|---|---|
| Method | Accuracy (%) | AUC | Accuracy (%) | AUC | Accuracy (%) | AUC |
| ori_Mesonet | 95.1 | 0.95 | 40.9 | 0.60 | 65.9 | 0.56 |
| v3_h | 92.5 | 0.99 | 48.8 | 0.62 | 75.9 | 0.77 |
| v3_v | 94.4 | 0.98 | 47.6 | 0.59 | 70.7 | 0.66 |
| v5 | 96.8 | 0.99 | 74.2 | 0.79 | 83.3 | 0.85 |
| v7_h | 96.6 | 0.99 | 63.7 | 0.73 | 60.1 | 0.74 |
| v7_v | 97.1 | 0.99 | 72.3 | 0.76 | 79.0 | 0.81 |
| v10 | 97.1 | 0.99 | 75.8 | 0.78 | 76.3 | 0.73 |
| v17 | 90.4 | 0.97 | 61.8 | 0.68 | 72.7 | 0.71 |
| v26 | 87.1 | 0.93 | 55.4 | 0.58 | 63.4 | 0.62 |
| v37 | 81.3 | 0.85 | 51.3 | 0.52 | 58.3 | 0.60 |

Table 5. Accuracy (%) of different segment methods.

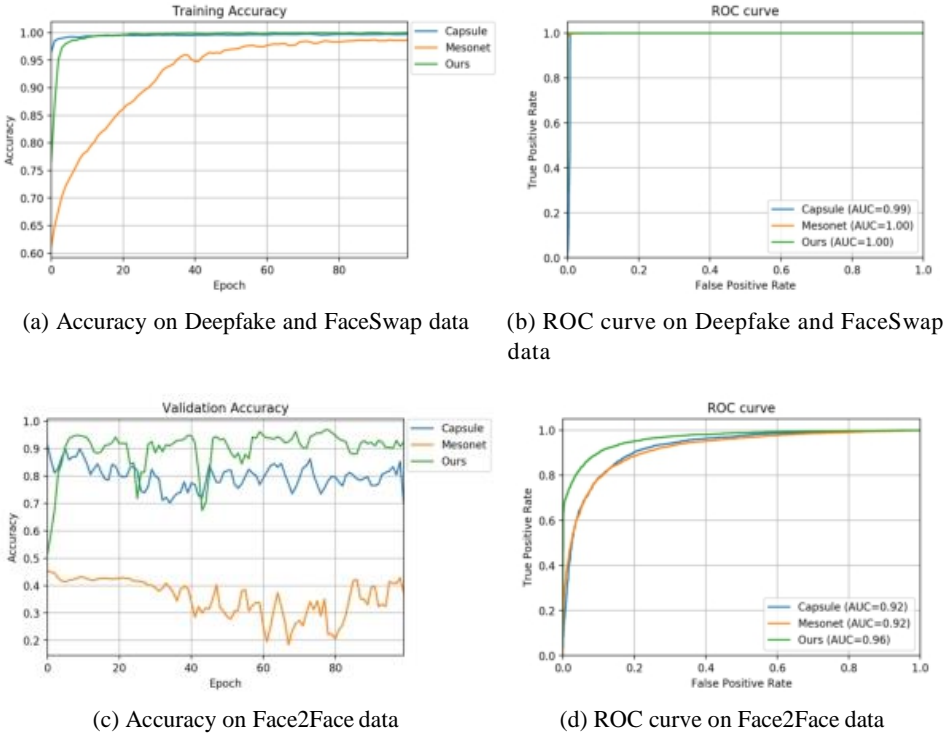

(a) Accuracy on Deepfake and FaceSwap data

(b) ROC curve on Deepfake and FaceSwap data

(c) Accuracy on Face2Face data

(d) ROC curve on Face2Face data

Fig. 13. Training on Deepfake and FaceSwap data and comparing Face2Face dataset accuracy with AUC results.

## 4.5 Compare Central Blocks of Different Shapes

This section compares the effect of central block shape on image recognition. Specifically, 10%, 20% to 90% of the center of the image was trained, as shown in Figure 18. The purpose of this was to see if the main feature was only at the center. As above, Mesonet's network architecture was used as a baseline. The accuracy and AUC values are shown in Table 6. It was observed that when the training data approached the whole picture, better results were obtained. The results also showed that training all latent features did not necessarily yield the best results, because there are slight differences between the image feature and the training data in different data sets. In Figure 19a and Figure 19b, the training data and the validation data are split by the same data set, resulting in better accuracy and AUC values. Figure 19c and Figure 19d show that when the proportion of training data in the central block is relatively low, the classification results of the proposed model will be worse because only a few features are captured. Based on the ROC curve, the model classifies all input images as either negative or positive, which results in the AUC value

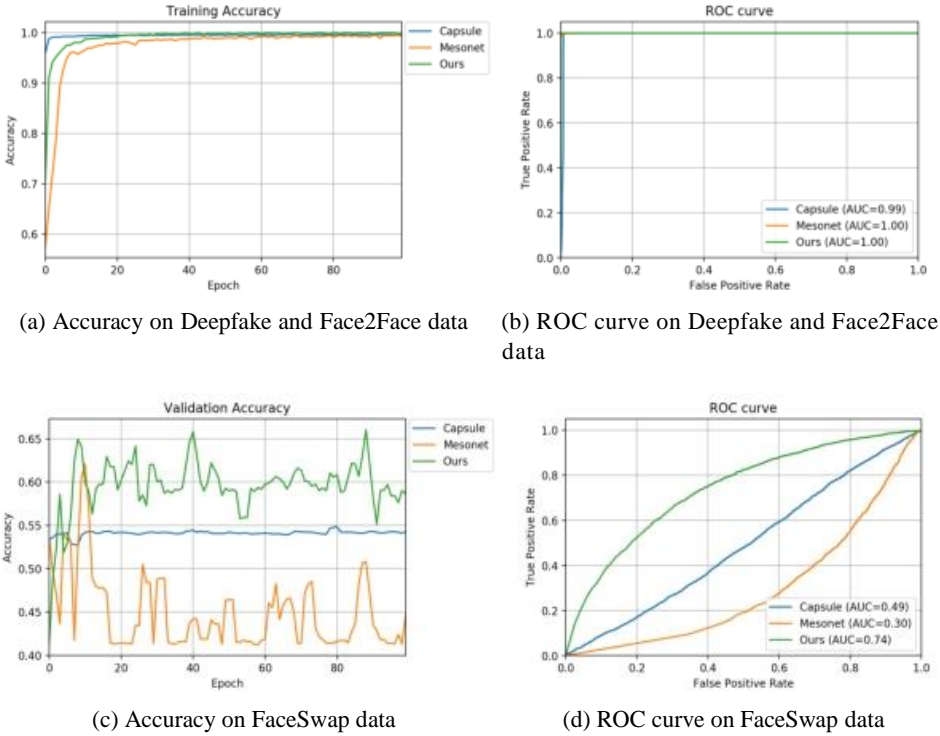

(a) Accuracy on Deepfake and Face2Face data

(b) ROC curve on Deepfake and Face2Face data

(c) Accuracy on FaceSwap data

(d) ROC curve on FaceSwap data

Fig. 14. Training on Deepfake and Face2Face data and comparing FaceSwap dataset accuracy with AUC results.

fluctuating and closing in on 0.55 when the range of the training data is 10% to 40% of the central block. When the training data was increased from 50% to 70% of the center block, the results of the model classification increased slightly, indicating that more image features were trained. When more latent features were trained, the classifier classified more forged features. Therefore, the accuracy and AUC values were higher. Specifically, a model training more main forged features achieves better classification results. We can see that the accuracy and AUC values continue to improve as the training data set increases from 70% to 90% of the central block. In this way, it can also explain why the proposed method chooses the v5 mode to cut the image. This result can also explain why the proposed method performs better than the v7_h and v7_v modes shown in Figure 16.

In summary, a comparison of different methods should not be based solely on accuracy. The results obtained by the ROC curve were more suitable for analyzing the generalization of the model. Moreover, cut-off point, as a new classification threshold, can improve the accuracy of the model. However, different model architectures can also use the proposed segmentation method because it allows the model to learn detailed functions. In this way, if the previous model is combined with our proposed image segmentation method, it can get a better image classification rate than the original model.

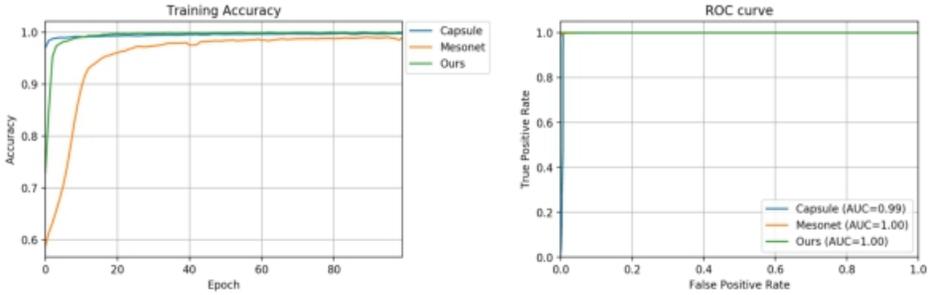

(a) Accuracy on FaceSwap and Face2Face data

(b) ROC curve on FaceSwap and Face2Face data

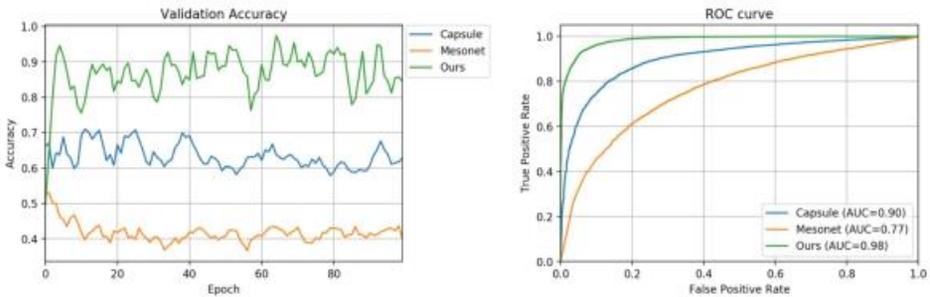

(c) Accuracy on Deepfake data

(d) ROC curve on Deepfake data

Fig. 15. Training on FaceSwap and Face2Face data and comparing Deepfake dataset accuracy with AUC results.

| Method | FF++ (vali) | | DeepFaceLab | | StyleGAN | |
|---|---|---|---|---|---|---|
| | Accuracy (%) | AUC | Accuracy (%) | AUC | Accuracy (%) | AUC |
| ori_Mesonet | 95.1 | 0.98 | 40.9 | 0.60 | 65.9 | 0.56 |
| cen10 | 84.7 | 0.92 | 39.0 | 0.53 | 50.2 | 0.51 |
| cen20 | 86.9 | 0.93 | 42.1 | 0.51 | 50.7 | 0.52 |
| cen30 | 86.1 | 0.94 | 41.3 | 0.57 | 52.2 | 0.53 |
| cen40 | 87.1 | 0.93 | 42.3 | 0.55 | 54.1 | 0.53 |
| cen50 | 89.3 | 0.96 | 42.7 | 0.56 | 53.5 | 0.55 |
| cen60 | 90.0 | 0.97 | 43.5 | 0.57 | 54.6 | 0.56 |
| cen70 | 91.1 | 0.96 | 43.6 | 0.58 | 57.3 | 0.55 |
| cen80 | 91.4 | 0.97 | 44.7 | 0.62 | 58.1 | 0.56 |
| cen90 | 92.3 | 0.97 | 44.5 | 0.62 | 61.2 | 0.61 |

Table 6. Accuracy (%) in central areas of different shapes.

## 5 CONCLUSION

The method proposed in this study is an ensemble model for detecting manipulated videos or images. This segmentation method exhibits improved accuracy and robustness. Moreover, the proposed method does not require too many additional training parameters. However, there is still

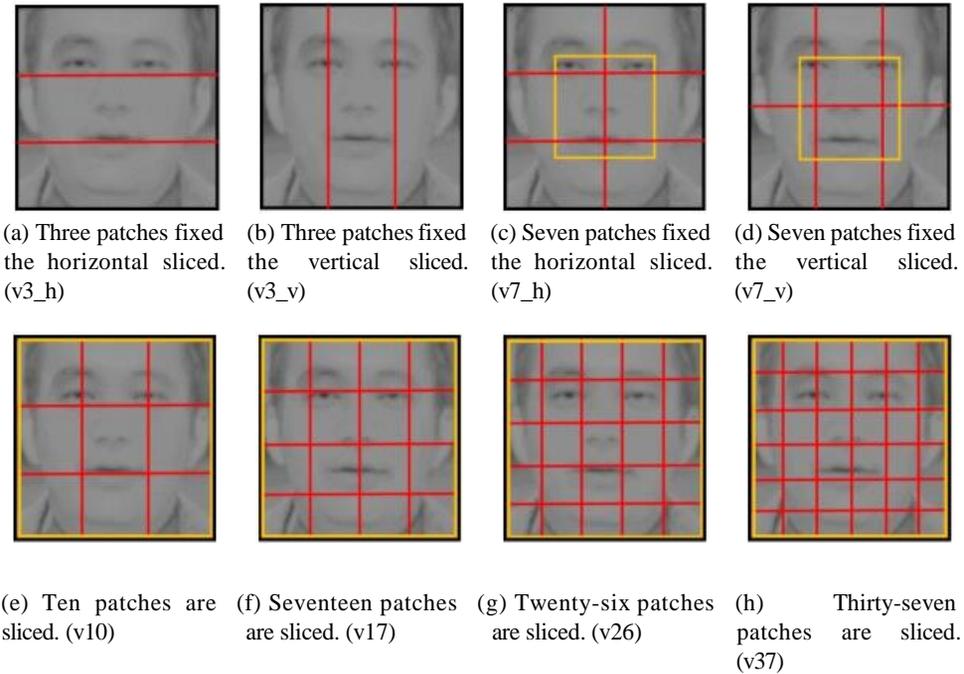

(a) Three patches fixed the horizontal sliced. (v3_h)

(b) Three patches fixed the vertical sliced. (v3_v)

(c) Seven patches fixed the horizontal sliced. (v7_h)

(d) Seven patches fixed the vertical sliced. (v7_v)

(e) Ten patches are sliced. (v10)

(f) Seventeen patches are sliced. (v17)

(g) Twenty-six patches are sliced. (v26)

(h) Thirty-seven patches are sliced. (v37)

Fig. 16. Eight different methods for image segmentations.

much room for improvement, and future work will aim to extend image forgery detection to areas of the human body other than the face, as well as improving the trained model's generalization ability.

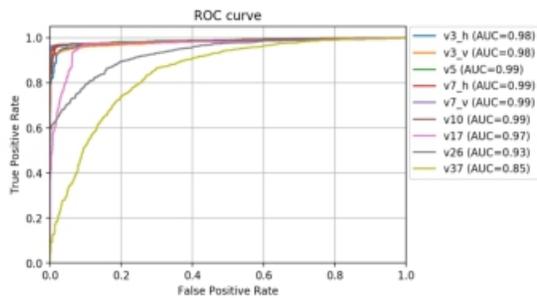

(a) ROC curve on FaceForensics++ (Validation)

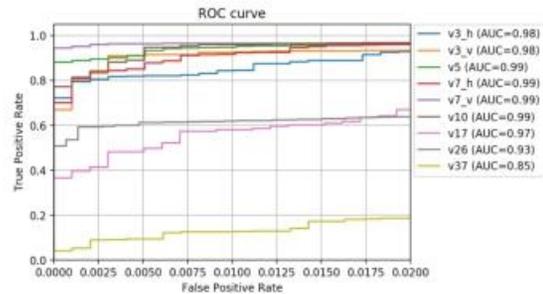

(b) ROC curve on FaceForensics++ (Validation)

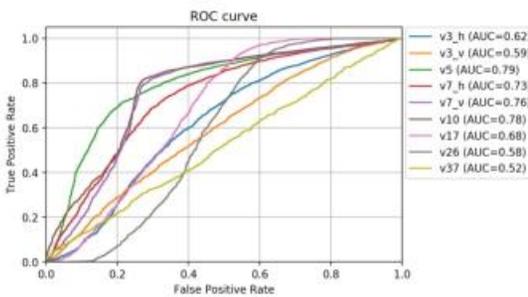

(c) ROC curve on DeepFaceLab

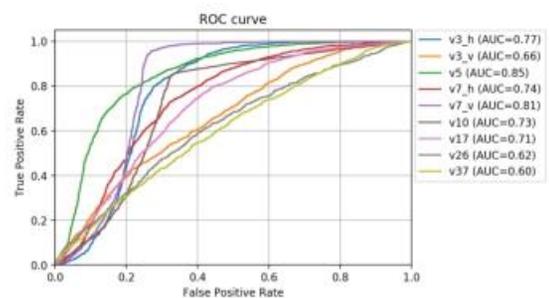

(d) ROC curve on StyleGAN

Fig. 17. Training on FaceForensics++ data and comparing the ROC curve of other datasets with AUC results.

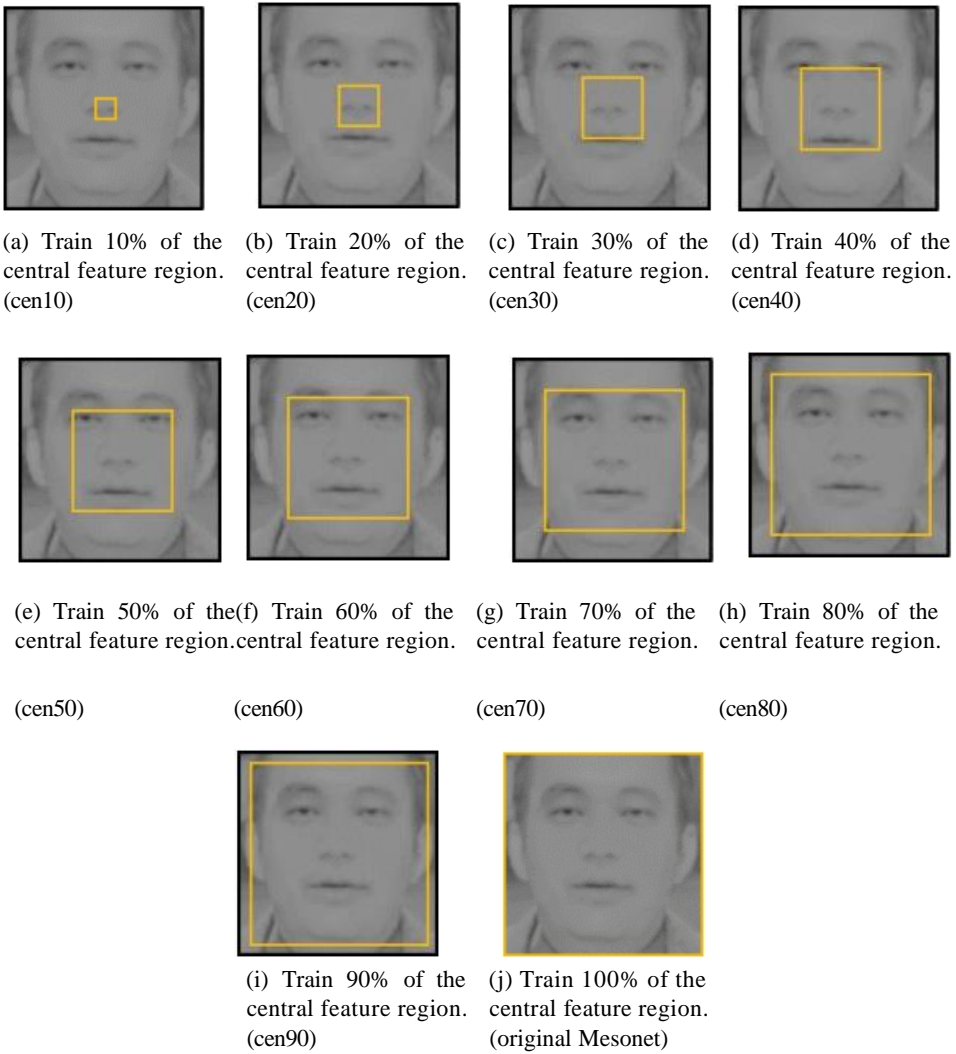

(a) Train 10% of the central feature region. (cen10)

(b) Train 20% of the central feature region. (cen20)

(c) Train 30% of the central feature region. (cen30)

(d) Train 40% of the central feature region. (cen40)

(e) Train 50% of the central feature region. (cen50)

(f) Train 60% of the central feature region. (cen60)

(g) Train 70% of the central feature region. (cen70)

(h) Train 80% of the central feature region. (cen80)

(i) Train 90% of the central feature region. (cen90)

(j) Train 100% of the central feature region. (original Mesonet)

Fig. 18. Central regions of different sizes.

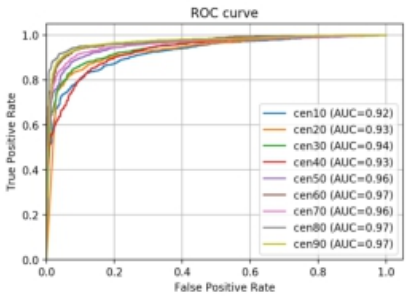 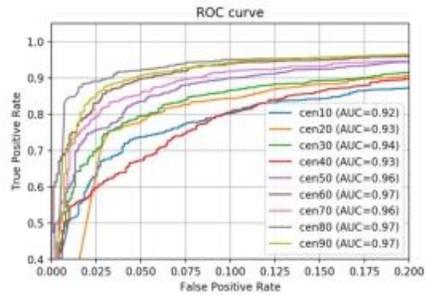

(a) ROC curve on FaceForensics++ (Valida-(b) ROC curve on FaceForensics++ (Validation)
tion)

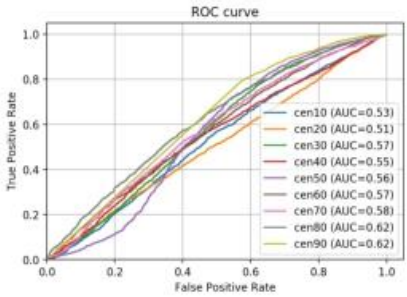 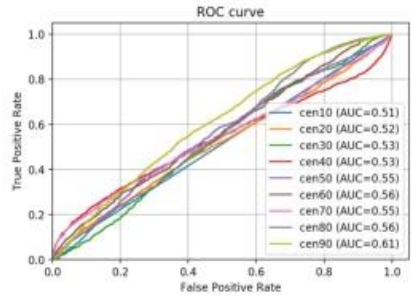

(c) ROC curve on DeepFaceLab (d) ROC curve on StyleGAN

Fig. 19. Training on FaceForensics++ data and comparing the accuracy of other datasets with AUC results.